\documentclass[runningheads]{llncs}
\usepackage{graphicx}
\usepackage[ruled,vlined]{algorithm2e}
\usepackage{amsmath}
\usepackage{hyperref}
\usepackage{longtable}
\usepackage{makecell, boldline}
\usepackage{multirow}
\usepackage{soul}
\usepackage{amssymb}
\usepackage{cite}
\begin{document}
\title{Precise Aerial Image Matching based on Deep Homography Estimation}
%
\author{Myeong-Seok Oh\inst{1} \and
Yong-Ju Lee\inst{2} \and
Seong-Whan Lee\inst{1,2,3}}
\authorrunning{M.-S, Oh et al.}
%
\institute{Department of Computer and Radio Communications Engineering
\and
Department of Brain and Cognitive Engineering
\and
Department of Artificial Intelligence\\
Korea University, Seoul, Republic of Korea\\
\email{\{ms\_oh, yz\_lee, sw.lee\}@korea.ac.kr}}
\maketitle              
\begin{abstract}
Aerial image registration or matching is a geometric process of aligning two aerial images captured in different environments. Estimating the precise transformation parameters is hindered by various environments such as time, weather, and viewpoints. The characteristics of the aerial images are mainly composed of a straight line owing to building and road. Therefore, the straight lines are distorted when estimating homography parameters directly between two images. In this paper, we propose a deep homography alignment network to precisely match two aerial images by progressively estimating the various transformation parameters. The proposed network is possible to train the matching network with a higher degree of freedom by progressively analyzing the transformation parameters. The precision matching performances have been increased by applying homography transformation. In addition, we introduce a method that can effectively learn the difficult-to-learn homography estimation network. Since there is no published learning data for aerial image registration, in this paper, a pair of images to which random homography transformation is applied within a certain range is used for learning. Hence, we could confirm that the deep homography alignment network shows high precision matching performance compared with conventional works.

\keywords{aerial image matching$\cdot$ image registration$\cdot$ homography \\transformation$\cdot$ geometric transformation}
\end{abstract}
\section{Introduction}
Recently, there is significant progress in many computer vision tasks~\cite{Yang_2007_PR, Roh_2007_PR, Bulthoff_2003_book, Lee_1997_NN}. Aerial image registration or matching, one of the computer vision tasks, is a geometric process of aligning two images taken in aerial. The two images captured in the same location are obtained in different environments, such as time, perspectives, and sensors. It is a pre-processing task before utilizing a variety of aerial image tasks. 
In the field of image match, deep learning-based methods~\cite{Wu_2013_SFM,Liao_2017_AAAI,Wu_2019_BE} have improved performance comparing to conventional approaches~\cite{Lowe04distinctiveimage,Bay_surf:speeded,Morel:2009:ANF:1658384.1658390}.
However, due to high resolution, large-scale transformation and various environmental characteristics, existing general image registration methods do not show well performance in aerial image registration.
Also, in previous studies of aerial image matching, similarity transformation or affine transformation has been mainly used. However, since the above transformation transforms each pixel homogeneously, it exhibits a partial disparity matching performance. This causes a fine matching error and degrades performance. To solve this problem, a transformation with higher degrees of freedom than similar transformation or affine transformation was adopted in the study of general image registration, and a higher matching rate was shown. However, applying a higher degree of freedom transformation to aerial photography makes it difficult to use a higher degree of freedom transformation that causes parallel or linear distortion, as properties such as flatness and linearity must be well maintained. Therefore, we describe how we can apply a high degree of freedom transformations even in aerial photography without compromising performance.
    \begin{figure}[t!]
		    \centering
			\includegraphics[width=9cm]{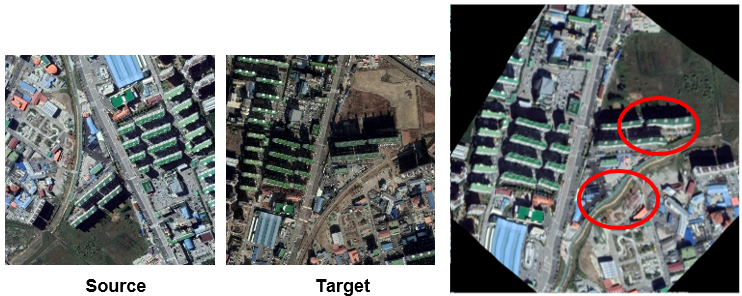}
			\caption{{The straight line characteristics of aerial images. Since most components of aerial image are building and road, there is a distortion of line when applying homography transformation directly.}}
			\label{fig:asymm} 
	\end{figure}  
To sum up, our contributions are~two-fold:
	\begin{itemize}
		\item We propose a novel method for estimating transformation parameters betweeb two aerial images. By progressively estimating the elements of whole parameters for affine, perspective, and homography transformations and training the matching network corresponding to each element, it is possible to successfully ensemble the regression parameters from each network and align a source image with target. 
		\item Our method enables the network to transform the aerial image without distortion of straight line. As an evaluation, we utilize probability of correct keypoints (PCK) metrics to access quantitative performance of alignment. Our method shows superior performance from the qualitative and quantitative assessment compared to the conventional methods.
		
	\end{itemize}

\section{Related Work}

As mentioned earlier, first looking at the recent research on general image matching~\cite{Brown:1992:SIR:146370.146374}, a method of applying parametric transformation (similar~\cite{Chopra_05_learninga}, affine~\cite{Morel:2009:ANF:1658384.1658390}, perspective~\cite{Song_2019_PR}) to complete rough registration and then applying non-parametric transformation (TPS~\cite{TPS_89}) to precisely match. These papers experiment with various conversion methods through various ablation studies and compare the accuracy of each transformation method. Looking at the results of this study, it can be seen that better performance comes out when a transformation with a high degree of freedom is applied.
However, if you look at the recent research on matching aerial images, it can be confirmed that similar transformation or affine transformation is still used in general. Therefore, a minute precision error occurs partially, and it is difficult to expect a sufficient precision matching performance. However, if we look at the paper that recently applied the similarity transformation and homography transformation to aerial image-stitching and compared the performance, it can be seen that the aerial image matching performance is rather low even when homography transformation is used. The paper's argument is that the higher the height of aerial images, the higher the performance of the similarity transformation that does not consider the perspective because the perspective caused by the height difference of the building becomes very small. However, even if the sense of perspective according to the height has been reduced, there are comparatively close and distant parts of the subject according to the shooting angle of the remote sensing image, and this includes perspective. Therefore, it is assumed that there is room for precise performance improvement if a higher degree of freedom transformation including perspective transformation is applied well in aerial image registration.
The affine and similarity transformations of previous studies also show meaningful results, but as tasks after registration expect more and more accurate matching performance, a higher degree of freedom transformation is applied to aerial images to further enhance performance. Therefore, we propose a structure that can stably learn the homography estimation network. By comparing the proposed learning structure with the learning structures of previous studies, the reason why homography estimation is difficult is revealed, and a method to solve the network learning problem is presented.

\section{Method}
\label{sec:proposed}
\begin{figure*}
		\centering
			\includegraphics[width=\textwidth]{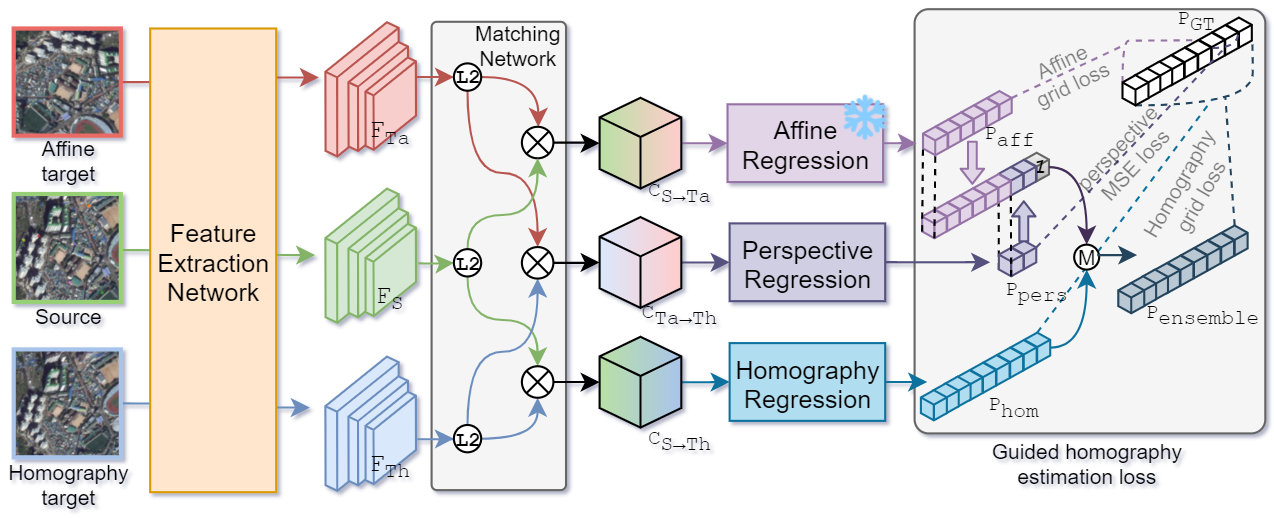}
			\caption{Overall of the proposed training process. The proposed learning network estimates a stable affine parameter from the previously learned affine regression network and at the same time estimates the perspective parameter to generate a parameter that guides the learning of the homography regression network. By learning the parameters estimated in the homography regression network, which is the goal of our learning, the ensemble with the above guide parameters, we can learn more stable and faster than learning homography estimation at once.}
			\label{fig:fig_2}
	\end{figure*}

    The proposed framework is robust to various changes in aerial images and improves precise consistency. The matching process for the proposed framework is as follows: (1) Target image random transformation preprocessing for data enhancement, (2) Estimate the affine parameters between the source image and the affine target image, (3) Estimate the perspective parameter between the affine target image and the homography target image, (4) Homography parameters are created by combining the estimated affine and perspective parameters, (5) The homography parameters between the source image and the homography target image are estimated and ensembled with the homography parameters generated in step (4), (6) The homography estimation network is stably learned while learning ensemble parameters. In~Figure~\ref{fig:fig_2}, we present the the proposed framework with training process.
	
	\subsection{Pre-processing data before training}
    As input to the network, a pair of images $ (I_S, I_ {Ta}, I_ {Th}) $ with different geometric properties in the same area are given. In figure \ref{fig:fig_3}, the image pairs consist of a target image converted from a source image by random homography and a target image converted into six parameters obtained by subtracting the perspective parameter from the homography transformation parameter. At this time, the target image$(I_{th})$ is randomly homography transformed to include geometric differences, and then injected into the network.
    \begin{figure}[h]
		\centering
		\includegraphics[width=9cm]{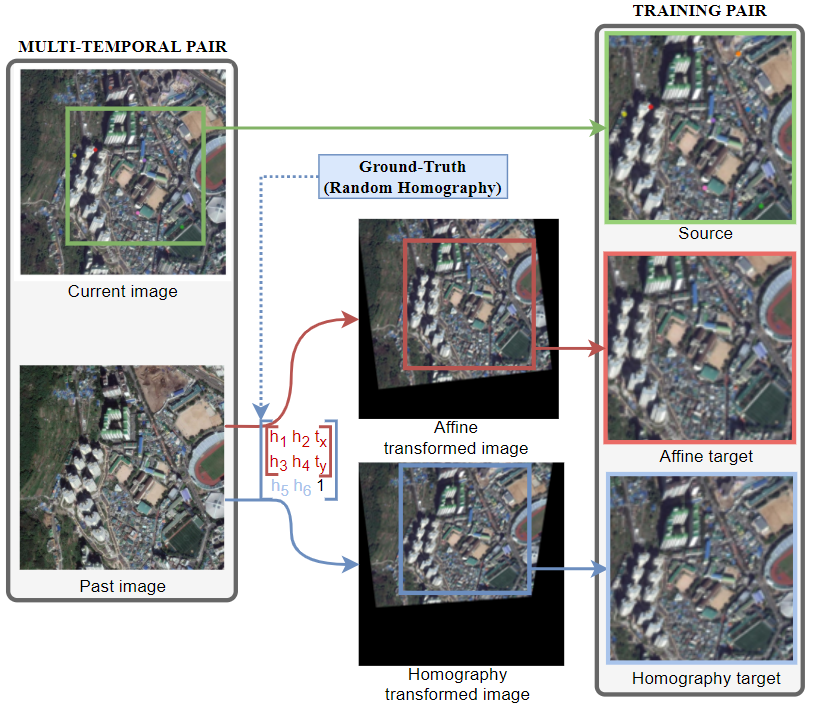}
		\caption{{Process of generating the training pairs.} Given a pair of current and past images sensed in the same area during training, an affine transformed image with 6 parameters(${[h_1, h_2, t_x, h_3, h_4, t_y]}$) among random homography parameters and a homography transformed image with eight parameters(${[h_1, h_2, t_x, h_3, h_4, t_y, h_5, h_6]}$) are obtained.}
		\label{fig:fig_3}
	\end{figure}
    A geometric feature that maintains parallelism is created between the source and the affine target, and a geometric feature that maintains the only linearity is created between the source and the homography target.In addition, the feature of geometric perspective transformation is given between the affine target and the homography target. This means that you can create a homography target by tilting the affine target back and forth. This way, you can create a dataset that can partially label homography parameters.A total of 8 homography parameters can be expressed as follows:
    \begin{equation}
    	\label{eq:eq_0}
    	[h_1, h_2, t_x, h_3, h_4, t_y, h_5, h_6],
    	\end{equation}
    where $h_1, h_2, h_3, h_4$ represent rotation transformation. Also, $h_1,h_4$ represent scale transformations, and $h_2,h_3$ represent left and right shear transformations $t_x, t_y$ represent parallel movements along the $x$ and $y$ axes. Finally, $h_5,h_6$ represent perspective transformations that represent the forward and backward tilt.

\unskip
    	
	\subsection{Transformation matrix regression network}\label{ssec:bi_direction}
\unskip
Basically, we adopt a method that mimics the stages of feature extraction, matching, transformation matrix regression, and transformation, which are traditional matching methods with deep learning.the features of each image are extracted. 
We extract feature maps $(f_S, f_T, f_{T'})\in \mathbb{R} ^{h'\times w'\times d'}$ from input images. Obtaining a feature map through the feature extraction network is expressed as follows:
    \begin{equation}
	\label{eq:eq_1}
	\mathcal{F}: \mathbb{R} ^{h\times w\times d}\rightarrow\mathbb{R} ^{h'\times w'\times d'},
	\end{equation}
	where $(h, w, d)$ denote the heights, widths, and~dimensions of the input images. $(h', w', d')$ are expressed as the height, width, and dimension of the extracted feature map.
	the transformation parameters is predicted in The regression step. The regression network $\mathcal{R}$ estimates the geometric transformation parameters directly when dense correspondence maps are passed through the network $\mathcal{R}$ as follows:
    \begin{equation}
	    \mathcal{R}: \mathbb{R} ^{h'\times w'\times (h'\times w')}\rightarrow\mathbb{R}^{DoF},
	\end{equation}
$DoF$ means the degrees of freedom of the transformation model.
	\begin{figure}[h]
		\centering
		\includegraphics[width=8cm]{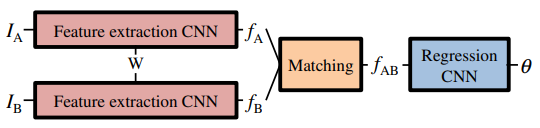}
		\caption{{Transformation matrix estimation network.}~\cite{Rocco_2017_CVPR}}
		\label{fig:fig_3}
	\end{figure}
    The extracted feature map shows the features of each pixel, and we make a correspondence map by calculating the similarity between the features of each pixel. When calculating the similarity, the cosine similarity was used by dot product of the generalized source and target feature maps. This matching network is expressed as follows:
\begin{align}
    		\label{eq:eq_2}
    		c_{S\rightarrow T}(i, j, k)&=\mathcal{C}(f_S(i_k, j_k), f_T(i,j))\nonumber\\
    		& = f_T(i,j)^{f}_{S}(i_k, j_k).
    	\end{align}

The dense correspondence map $c_{S\rightarrow T}$ aligns the source feature map $f_S$ with the target feature map $f_T$. similarity score between two points is computed by passing through $c_{S\rightarrow T}$.

The affine transform does not cause significant distortion because it maintains parallel components in the image. In addition, homography conversion can convert pixels more flexibly with a higher degree of freedom than affine conversion, thereby increasing the fine matching rate. We take advantage of this characteristic to ensemble the final transformation matrix. This transformation matrix is robust to linear distortion and increases the fine matching rate of aerial image registration with more flexible coordinate movement.
		We describe the characteristics of each transformation and move on to make good use of the advantages of affine transformation and homogeneous transformation. First, each transformation matrix can be expressed as follows:
\begin{equation}
		\label{eq:eq_4}
    		[a_1, a_2, t_x, a_3, a_4, t_y]\Longrightarrow 
    		\begin{bmatrix}
    		a_1 & a_2 & t_x \\
    		a_3 & a_4 & t_y
    		\end{bmatrix},
		\end{equation}
		\begin{equation}
		\label{eq:eq_5}
    		[h_1, h_2, t_x, h_3, h_4, t_y, h_5, h_6, 1]\Longrightarrow 
    		\begin{bmatrix}
    		h_1 & h_2 & t_x \\
    		h_3 & h_4 & t_y \\
    		h_5 & h_6 & 1
    		\end{bmatrix}.
		\end{equation}

		In the affine transformation parameters $[a_1, a_2, t_x, a_3, a_4, t_y]$, $a_1\sim a_4$ represent the scale, rotated angle and tilted angle, and~$(t_x, t_y)$ denotes the $(x$-axis, $y$-axis$)$ translation.
		Applying the estimated homography transformation matrix to the image follows the formula:
	\begin{equation}
		\begin{aligned}
		    \begin{bmatrix}
    		x'\\
    		y'\\
    		z'
    		\end{bmatrix}
    		=
		    \begin{bmatrix}
    		h_1 & h_2 & t_x\\
    		h_3 & h_4 & t_y\\
    		h_5 & h_6 & 1
    		
    		\end{bmatrix}
    		\bullet
    		\begin{bmatrix}
    		x \\
    		y \\
    		1
    		\end{bmatrix},\\
    		\begin{bmatrix}
    		x'\\
    		y'\\
    		1
    		\end{bmatrix}
    		=
    		\begin{bmatrix}
    		x'\\
    		y'\\
    		z'
    		\end{bmatrix}
    		\times {1 \over z'},\\
    		x'={h_1\cdot x+h_2\cdot y+t_x \over h_5\cdot x+h_6\cdot y+1},\\
    		y'={h_3\cdot x+h_4\cdot y+t_y \over h_5\cdot x+h_6\cdot y+1},
    	\end{aligned}
		    \label{eq:apply affine transformation}
	\end{equation}
	where $(x',y')$ is the coordinates of the transformed pixel. The transformed image is obtained by calculating the coordinates of all pixel$(x,y)$s in the source image. At this time, $z'$ is divided to express the coordinates in homogeneous coordinates form.
	
\subsection{Loss function for deep homography alignment}
As mentioned in the introduction, most aerial image registration studies use similarity transforms or affine transforms. This is to ensure that objects in the aerial image are not distorted. Similarity transforms and affine transforms with relatively low degrees of freedom do not cause significant distortion because parallelism or angles between line segments are maintained during the transform. It is also easy to learn because it estimates a small number of parameters.
After pre-learning this, we design the entire network structure like figure~\ref{fig:fig_2}, fix the parameters of the pre-trained affine estimation block, and learn the remaining parameters. The structure of such a network helps the learning of the homography estimation network and enables faster learning. In addition, if the estimated homography causes severe perspective distortion, the distortion can be reduced by the affine parameter.
	
As the baseline loss function, we adopt the transformed grid loss~\cite{Rocco_2017_CVPR} in the training procedure. Given the predicted transformation $\hat{\theta}$ and the ground-truth $\theta^{gt}$, the~baseline loss function $l(\hat{\theta},\theta^{gt})$ is obtained by the following:
\begin{equation}
\label{eq:loss}
    \begin{aligned}
		l(\hat\theta_{aff},\hat\theta_{pers},\hat\theta_{hom},\hat\theta_{en},\theta^{gt}_{aff},\theta^{gt}_{pers},\theta^{gt}_{hom})\\=\alpha\cdot\ell(aff)+\beta\cdot \ell(pers)+\gamma\cdot\ell(hom)+\delta\cdot\ell(en),
	\end{aligned}
\end{equation}
		where $\alpha,\beta,\gamma,\delta$ are hyper parameters for weighted summation. $\ell(aff), \ell(hom), \ell(en)$ are transformation grid losses. The grid loss is defined following:
	\begin{equation}
		\ell(aff)=\frac{1}{N}\sum_{i,j=1}^Nd(\mathcal{T}_{\hat\theta_{aff}}(x_i, y_j),\mathcal{T}_{\theta^{gt}_{aff}}(x_i, y_j))^2,
    \end{equation}
    \begin{equation}
		\ell(hom)=\frac{1}{N}\sum_{i,j=1}^Nd(\mathcal{T}_{\hat\theta_{hom}}(x_i,
		y_j),\mathcal{T}_{\theta^{gt}_{hom}}(x_i, y_j))^2,
    \end{equation}
    \begin{equation}
		\ell(en)=\frac{1}{N}\sum_{i,j=1}^Nd(\mathcal{T}_{\hat\theta_{en}}(x_i, y_j),\mathcal{T}_{\theta^{gt}_{hom}}(x_i, y_j))^2,
    \end{equation}
		
where $N$ is the number of grid points, $\mathcal{T}_{\hat{\theta}}(*)$ and $\mathcal{T}_{\theta^{gt}}(*)$ are the transforming operations parameterized by $\hat{\theta}$ and $\theta^{gt}$, respectively.
In equation \ref{eq:loss}, $\ell(pers)$ represents the perspective loss, and the difference between the two estimated perspective parameters and the perspective parameter of the ground truth is calculated through MSE. This is because the image cannot be converted with this parameter.

  \begin{figure}[t!]
	\centering
	\includegraphics[width=\textwidth]{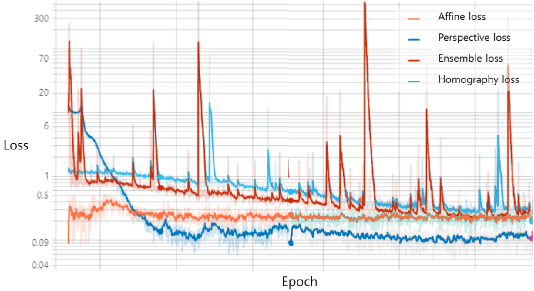}
	\caption{{Learning curve of the proposed model.} In the graph, the x-axis represents the learning time and the y-axis represents the loss value.}
	\label{fig:fig_5}
    \end{figure}
    
In our experiment, we set these parameters to (0.3, 0.4, 0.1, 0.2), respectively. The loss function for teaching is calculated from the source image, the affine target image, the homography target image, the homography parameter, the affine parameters up to the 6th and the perspective parameters of the 7th and 8th. Therefore, the loss function weights the affine loss, perspective loss, ensemble loss, and homography loss. The grid loss described above was used for the affine, ensemble and homography loss that can be geometrically transformed, and mse is used for the perspective loss. Looking at the learning pattern in Figure~\ref{fig:fig_5}, the perspective loss and the ensemble loss tend to decrease first, while the homography loss decreases and the learning is performed. Since affine and perspective loss play an important role in the overall learning, a weight of 0.3, 0.4 was given to each, and a weight of 0.1, 0.2 was given to homography and ensemble loss. This was tested in a heuristic way and it is worth trying various attempts for optimization.
\begin{figure*}
    \centering
    \includegraphics[width=11cm]{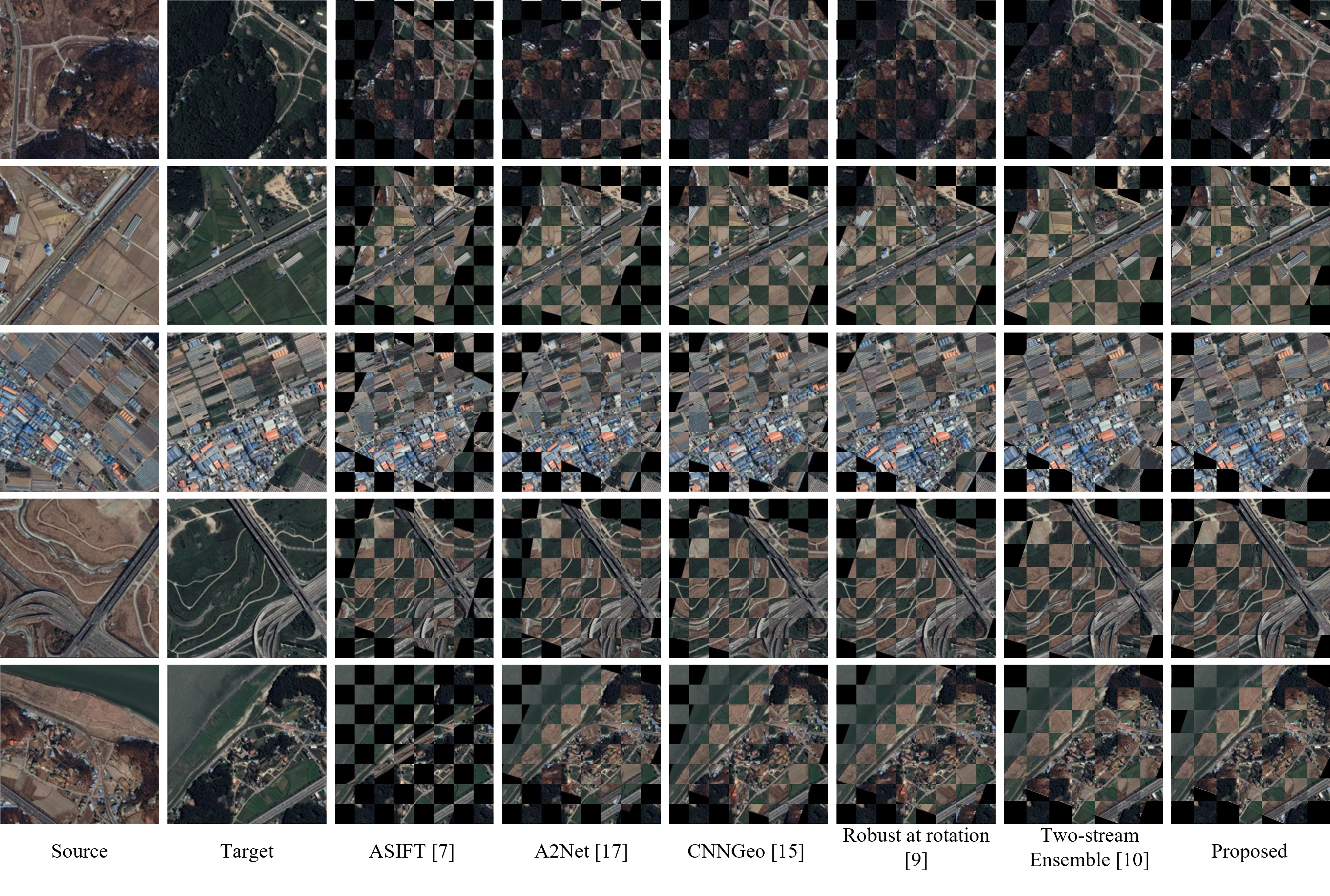}
    \caption{The two images from the left are the input and target images, respectively. Afterwards, the matching result of each methodology is shown on a checkerboard. The proposed method can confirm that the break or distortion of the road is minimized.}
    \label{fig:fig_4}
\end{figure*}

\section{Experiment}
\label{sec:experiments}
	In this section, we present the implementation details, experimental setup and results of the proposed network. In addition, quantitative and qualitative evaluations are conducted for the performance of the proposed matching network and compared with previous studies. In addition, by comparing the learning performance of the proposed learning method in the abstinence study section, we prove that the learning structure of the proposed network enables stable learning of the network that estimates difficult parameters.

	\subsection{Implementation~Details}
	When creating the training data, we have included a rotation of 180 degrees, an inclination of 60 degrees, a perspective tilt within 20 degrees, and a vertical and horizontal movement within 100 pixels. In addition, the homography regression network was modified to increase the complexity by adding a two-stage hidden layer to better estimate more parameters.

	\subsection{Experimental~Setup}

	\subsubsection{Training}
    The aerial image used for training was extracted from Google Earth, and a random homography transformation parameter within a certain range was generated using a multi-temporal aerial image pair, and the source image was partially used with 6 affine parameters and 8 homography parameters. A pair of labeled training data are defined by creating an affine target and a homography target.
    
	\subsubsection{Evaluation}
	To demonstrate quantitative result of the proposed method, we use PCK~\cite{pck}. PCK is the ratio of correctly converted keypoints among all keypoints between two images and is defined as follows: \begin{equation}
		\label{eq:eq_10}
		PCK=\frac{\sum_{i=1}^n\sum_{p_i}1[d(\mathcal{T}_{\hat{\theta}}(p_i),\mathcal{T}_{\theta^{gt}}(p_i))<\tau\cdot\max(h,w)]}{\sum_{i=1}^n|p_i|},
	\end{equation}
	where $p_i$ is the $i_{th}$ point, which consists of $(x_i,y_i)$, and~$\tau\cdot\max(h,w)$ refers to the tolerance term in the image size of $h\times w$. Intuitively,  In order to calculate the PCK, we need a test dataset including ground truth. In order to fairly evaluate our model for self-supervised learning, we need different training and test sets. So, we create a test set for PCK like this: First, features are extracted using SIFT~\cite{Lowe04distinctiveimage} from aerial image pairs in different time zones having the same center coordinates, and 20 coordinates that are viewed as the same point are constructed. And the coordinates converted from 20 coordinates are constructed using a larger range of random homography parameters than when creating the training data set. The test data set thus created contains more diverse geometric changes than the data set when training the network. In addition, by not using the aerial image used for training at this time, the separation of the training data set and the test set was successfully carried out. Therefore, quantitative evaluation is carried out based on how close the feature points extracted from the proposed network, the correspondence pair used for matching, and the location where the points are converted are close to the 20 coordinates of the ground truth.

\begin{table}[h]
\caption{Quantitative results. Comparisons of probability of correct keypoints (PCK) in the aerial images. CNNGeo~\cite{Rocco_2017_CVPR} is evaluated the pre-trained model by general images and the fine-tuned model by the aerial images. Both CNNGeo models use feature extractor of ResNet101.}
\centering
\resizebox{\columnwidth}{!}{
	\begin{tabular}{lccc}
		\hline
		\multirow{2}{*}{\textbf{Methods}} & \multicolumn{3}{c}{\textbf{PCK (\%)}} \\
		                         & \boldmath{$\tau=0.05$} 
		                         &  \boldmath{$\tau=0.03$}
		                         &  \boldmath{$\tau=0.01$} \\
		\hline
		SURF~\cite{Bay_surf:speeded} & 21.7 & 18.1 & 15.3 \\
		SIFT~\cite{Lowe04distinctiveimage} & 31.2 & 25.9 & 13.7 \\
		ASIFT~\cite{Morel:2009:ANF:1658384.1658390} & 34.8 & 27.9 & 17.9 \\
		OA-Match~\cite{Song_2019_PR} & 34.9 & 27.8 & 18.2 \\
		\hline
		CNNGeo~\cite{Rocco_2017_CVPR}  (pretrained) & 17.8 & 10.7 & 2.5 \\
		CNNGeo (fine-tuned) & 43.3 & 11.4 & 7.6\\
		Two-stream ensemble~\cite{Park_2020_remote}; SE-ResNeXt101~\cite{Hu_2018_CVPR} &{49.9} & {37.1} & {11.3}\\
		\hline
		Proposed~ & {\bf87.8} & {\bf60.2} & {\bf27.8}\\
		\hline
	\end{tabular}
}
\label{tab:tab_1}
\end{table}

\subsection{Quantitative~results}
\label{sssec:quantitative}
Table~\ref{tab:tab_1} compares PCK on aerial image data with large transformation between the conventional computer vision methods~\cite{Lowe04distinctiveimage, Bay_surf:speeded, Morel:2009:ANF:1658384.1658390, Fischler:1981:RSC:358669.358692, Song_2019_PR} and deep learning-based methods~\cite{Rocco_2017_CVPR,Rocco_2018_CVPR,Park_2020_remote}. There are quite a number of critical failures globally in Conventional computer vision methods. In addition, deep learning-based methods also show excellent performance, but when a dataset containing a large perspective change is applied, it shows insufficient performance.
		
	\begin{figure}[t!]
		\centering
		\includegraphics[width=\textwidth]{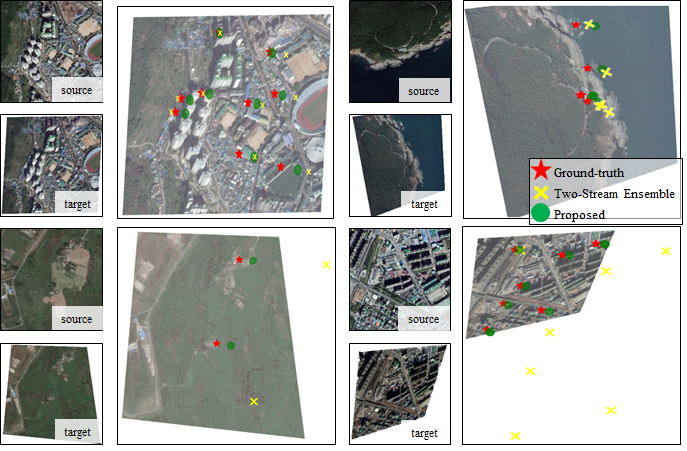}
		\caption{{Comparison of fine matching rates through key points visualization.} If the perspective transformation is not included as much as in the upper left, the baseline model also shows a high matching rate. However, if there is a perspective transformation, it can be confirmed that it shows extreme errors or that the proposed model matches more accurately.}
		\label{fig:fig_8}
	\end{figure}

\subsection{Qualitative~results}
\begin{figure*}
		\centering
		\includegraphics[width=11cm]{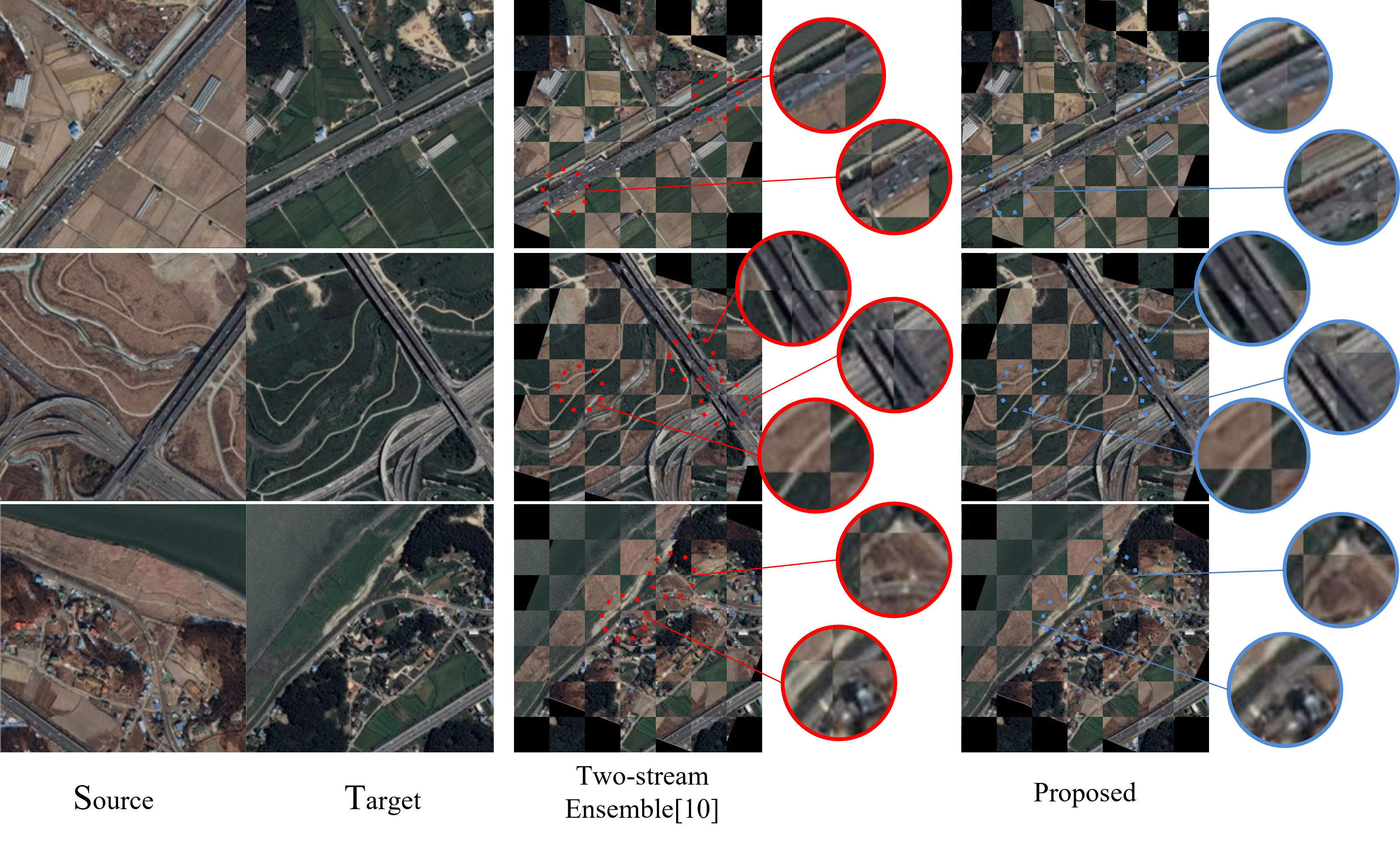}
		\caption{{Overlay the registration result on a checkerboard.} The proposed method shows accurate matching performance while minimizing the breakage or distortion of the road seen in the scene.}
		\label{fig:fig_7}
	\end{figure*}

We show some image matching results on aerial images in Figure~\ref{fig:fig_4}.
In Figure~\ref{fig:fig_8}, Precise matching ratio comparison was conducted through visualization of key points. It was confirmed that the ground-truth key points and the key points of the proposed method were closer.In~Figure~\ref{fig:fig_7}, If you look closely at the result of overlaying the registration result on the checkerboard, you can see that the linear distortion of the proposed model is less. The proposed method was confirmed to be more sophisticated than the comparable method.
	\unskip
\subsection{Comparison of training performance}
We propose a method of learning a network that estimates more difficult parameters by using the knowledge of the previously learned parameter estimation network. First, if the pre-trained affine estimation network is frozen and the learning is conducted, the perspective regression network learning two parameters is more easily learned. Accordingly, as the perspective loss decreases, the ensemble loss using the affine parameter and the perspective parameter decreases, and the target homography estimation network is learned in this manner. At this time, if the pre-trained affine regression network and perspective regression network are frozen, more stable homography estimation network learning is possible. This learning method can be used in a number of fields that can partially learn networks that are difficult to learn.

\section{Limitation and Future Work}
Since there is no public training data for aerial image registration, it is difficult to perform a fair evaluation because a test dataset that was randomly generated in a similar manner to the training was used. However, we tested the accuracy of the network by applying a different range of random homography transformations than in training. In the future, a method that can apply a transformation with a higher degree of freedom than homography to aerial images should be studied, and this will be a very challenging task due to the characteristics of aerial images where maintaining linearity is very important. In addition, it is necessary to study how to find more accurate correspondence by fusion of features based on deep learning and the advantages of handcrafted features.

\section{Conclusion}
We propose a deep learning-based network that can precisely match aerial images. This network adopts a homography transformation method that can be well applied to the characteristics of aerial images that require minimal distortion. This allows a more precise matching result. Also, estimating homography transformations including perspective transformations is quite difficult because even small angular changes lead to large perspective distortions. The proposed network first applies affine transformation to feed a more generalized image to the homography estimation network, so homography learning becomes possible. Future research should be conducted in the direction of minimizing the distortion of the object in the angular image and reducing the fine registration error. Therefore, it is necessary to apply a higher degree of freedom, but it is difficult to learn to estimate a transformation of a higher degree of freedom than homography, and to deal with object distortion is difficult.

%
%
%
%

\end{document}